\documentclass[pdflatex,sn-mathphys-num]{sn-jnl}

\usepackage{graphicx}%
\usepackage{multirow}%
\usepackage{amsmath,amssymb,amsfonts}%
\usepackage{amsthm}%
\usepackage{mathrsfs}%
\usepackage[title]{appendix}%
\usepackage{xcolor}%
\usepackage{textcomp}%
\usepackage{manyfoot}%
\usepackage{booktabs}%
\usepackage{algorithm}%
\usepackage{algorithmicx}%
\usepackage{algpseudocode}%
\usepackage{listings}%

\theoremstyle{thmstyleone}%
\theoremstyle{thmstyletwo}%

\theoremstyle{thmstylethree}%

\begin{document}

\title[Article Title]{Vietnamese AI Generated Text Detection}

\author[1,2]{\fnm{Quang-Dan} \sur{Tran}}\email{21521917@gm.uit.edu.vn}

\author[1,2]{\fnm{Van-Quan} \sur{Nguyen}}\email{21521333@gm.uit.edu.vn}

\author[1,2]{\fnm{Quang-Huy} \sur{Pham}}\email{21522163@gm.uit.edu.vn}

\author[1,2]{\fnm{K. B. Thang} \sur{Nguyen}}\email{21521432@gm.uit.edu.vn}

\author*[1,2]{\fnm{Trong-Hop} \sur{Do}}\email{hopdt@uit.edu.vn}

\affil[1]{\orgdiv{Faculty of Information Science and Engineering}, \orgname{University of Information Technology}}

\affil[2]{\state{Vietnam National University}, \country{Ho Chi Minh City, Vietnam}}


\abstract{In recent years, Large Language Models (LLMs) have become integrated into our daily lives, serving as invaluable assistants in completing tasks. Widely embraced by users, the abuse of LLMs is inevitable, particularly in using them to generate text content for various purposes, leading to difficulties in distinguishing between text generated by LLMs and that written by humans. In this study, we present a dataset named \textbf{ViDetect}, comprising 6.800 samples of Vietnamese essay, with 3.400 samples authored by humans and the remainder generated by LLMs, serving the purpose of detecting text generated by AI. We conducted evaluations using state-of-the-art methods, including ViT5, BartPho, PhoBERT, mDeberta V3, and mBERT. These results contribute not only to the growing body of research on detecting text generated by AI but also demonstrate the adaptability and effectiveness of different methods in the Vietnamese language context. This research lays the foundation for future advancements in AI-generated text detection and provides valuable insights for researchers in the field of natural language processing.}




\maketitle

\section{Introduction}\label{sec1}

In recent years, the natural language processing (NLP) community has witnessed a paradigm shift with the introduction of Large Language Models (LLMs) \cite{devlin2018bert},\cite{yang2019xlnet}. The formidable capabilities of LLMs have raised concerns about distinguishing between their generated texts and human-written content. These concerns stem from two main issues. Firstly, LLMs are prone to producing fabricated or outdated information, potentially leading to the spread of misinformation and plagiarism. Secondly, there is a risk of malicious use, including disinformation dissemination, online fraud, social media spam, and academic dishonesty by students using LLMs for essay writing. To address these concerns, it is essential to implement measures that enhance the accountability and accuracy of LLMs, ensuring responsible use across various domains.
This detection task involves determining whether a given piece of text is generated by an LLM, essentially constituting a binary classification challenge.
Recent advancements in detector techniques, driven by innovations in watermarking, zero-shot methods, fine-tuning language models, adversarial learning, LLMs as detectors, and human-assisted methods, have significantly enhanced the capabilities of LLM-generated text detection.

In this study, we introduce a dataset called \textbf{ViDetect}, a pioneering publicly available benchmark Detect AI Generated Text dataset for Vietnamese essay. Our main contributions are described as follow:
\begin{enumerate}
    \item Constructing ViDetect dataset: a first benchmark Detect AI Generated Text dataset on Vietnamese. ViDetect refer to the essays. This dataset consists of nearly 6.800 samples that have undergone rigorous quality control measures, ensuring the highest quality of the dataset.
    
    \item Conducting various experiments employing several state-of-the-art language models including ViT5, BARTpho, mDerbertav3, PhoBERT, and Bert-multilingual on the Vi-Detect AI Generated Text dataset. These models have been fine-tuned and evaluated to investigate their effectiveness for the task.
    
    \item Undertaking a comprehensive analysis of the limitations and challenges encountered during the development of the ViDetect dataset, providing valuable insights to guide future research endeavors.

\end{enumerate}

\begin{figure*}[!h]
  \centering
  \includegraphics[width=01.\textwidth]{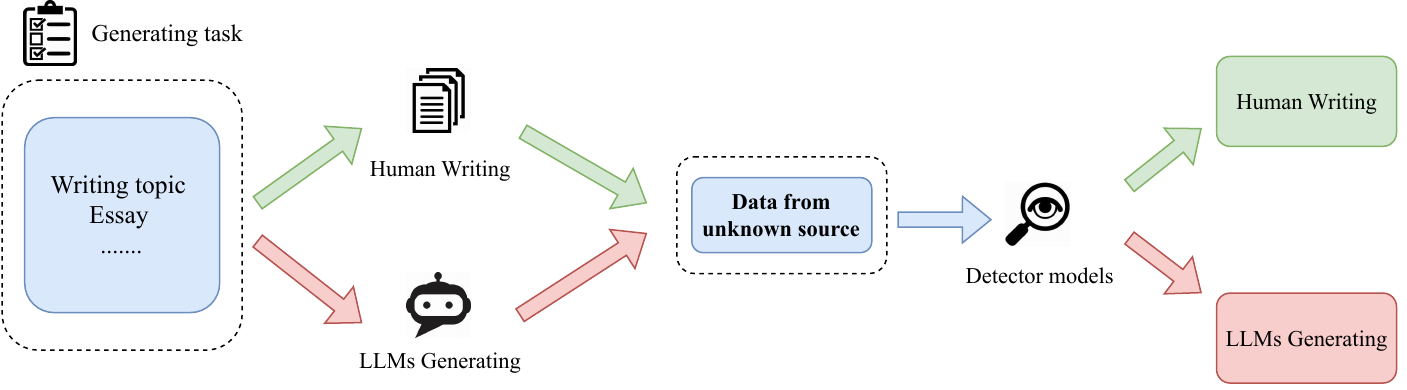}
  \caption{Picture of LLM-generated text detection task. This task is a binary classification task that detects whether the provided text is generated by LLMs
or written by humans.}
  \label{fig:model}
\end{figure*}

\section{Related Work}\label{sec2}
\subsection{Former LLM-Generated Text Detection
Datasets}\label{sec2.1}

High-quality datasets have a key role in advancing research in LLM-generated text detection. These datasets allow researchers to quickly develop and calibrate effective detectors and establish standardized metrics to evaluate the effectiveness of their methods. In this section, we introduce popular datasets used for LLM-generated text detection, widely contributed by recent studies.

\textbf{Human ChatGPT Comparison Corpus (HC3)}: The Human ChatGPT Comparison Corpus (HC3) \cite{guo2023close} is an early initiative designed to compare text generated by ChatGPT with human-written text. It comprises two datasets: HC3-en in English and HC3-zh in Chinese. HC3-en contains 58,000 human responses and 26,000 ChatGPT responses, HC3-zh consists of 22,000 human answers and 17,000 ChatGPT responses. However, it's important to note some limitations of the HC3 dataset, such as the lack of diversity in bootstrapped generated instructions.

\textbf{CHEAT}: The CHEAT dataset \cite{yu2023cheat} is designed to identify artificially generated misleading academic content produced by ChatGPT. The dataset consists of human-created academic abstracts sourced from IEEE Xplore. The dataset includes 15k human-written abstracts and 35k summaries crafted by ChatGPT. The limitation of the CHEAT dataset is its focus on specific academic disciplines, neglecting challenges that span different domains due to constraints in its primary data source.

\textbf{TweepFake}: The TweepFake dataset \cite{fagni2021tweepfake} is a foundational dataset designed for the examination of fake tweets on Twitter, sourced from both authentic and deceptive accounts. It compiles a total of 25,000 tweets, evenly divided between human-authored and machine-generated content. The machine-generated tweets were produced using a variety of methods, including GPT-2, RNN, Markov, LSTM, and CharRNN. While TweepFake remains a preferred dataset for numerous researchers, those utilizing LLMs should thoroughly evaluate its pertinence and robustness given the continuous advancements in technological capabilities.

\subsection{LLM-Generated Text Detection Methods}
\subsubsection{Watermark Technology}

Watermarking techniques is an indispensable part of detecting images generated by AI, contributing to protecting intellectual property rights and ownership in visual arts. With the emergence of LLMs, watermarking techniques has been extended to be applicable to identifying text generated by these models. Methods applying watermarking techniques include tatistical Methods \cite{kirchenbauer2023watermark}, Secret Key-Based Watermarking Technology \cite{zhao2022distillation}, etc.

\subsubsection{Zero-shot Methods}
The zero-shot technique is used to identify text generated by LLMs, aiming to establish detectors without the need for additional training through supervised signals. This approach assumes access to LLMs, where their inputs are evaluated based on specific features and statistical metrics. The pioneering work in using the zero-shot technique for detecting AI-generated text was by Corston-Oliver et al. \cite{corston2001machine}, paving the way for further research based on the zero-shot technique \cite{arase2013machine}.

\subsubsection{Fine-tuning LMs Methods} \label{fine-tuning}
We delve into the methods involved in fine-tuning a Transformer-based Language Model (LM) to distinguish between input text generated by LLM and text that is not. This strategy requires paired samples to facilitate supervised training. Pre-trained LMs have been shown to have strong natural language understanding capabilities, playing a key role in enhancing various tasks in Natural Language Processing (NLP), among them Text categorization is especially noteworthy. This method has been researched by the community and has shown good results \cite{li2023deepfake} \cite{fagni2021tweepfake}.

\subsubsection{Adversarial Learning Methods}

Adversarial learning methods related to fine-tuning LMs methods, notably the work of Koike et al. \cite{koike2023outfox}, reveal that adversarial training can be conducted without fine-tuning the model, with the context acting as guidance for a fixed-parameter model. The two main approaches of this method include enhancing adversarial samples to reinforce the robustness of the detector \cite{shi2024red}, and configuring both the attack model and the detection model simultaneously to create a confrontation between the two models, enhancing detection capability \cite{goodfellow2020generative}.

\subsubsection{LLMs as Detector}

Some studies have investigated the potential use of LLMs as detection tools to distinguish between texts generated by themselves or by other LLMs \cite{bhattacharjee2024fighting}. However, the studies have illustrated a decreasing reliability in using LLMs directly for detecting self-generated texts, particularly when compared to neural network and statistical methods. This trend becomes more evident when facing the increasing complexity of LLMs.

\subsubsection{Human-assisted Methods}

The human-assisted methods use to detect text generated by LLMs are a result of leveraging prior knowledge and analytical skills, providing significant interpretability and reliability in the detection process. Previous studies have extensively analyzed the imperfect writing abilities of LLMs compared to humans, such as inconsistencies in semantics and logic within sentences \cite{uchendu2023does}, as well as enhancing user recognition through various means \cite{gehrmann2019gltr}.

\section{Dataset}\label{sec3}

Currently, in Vietnam, there is no specifically created dataset to serve the task of AI text detection. Therefore, we have decided to construct a suitable dataset named \textbf{ViDetect} to support this endeavor. The process of developing our dataset is carried out meticulously and with high quality to ensure that it meets all the requirements and challenges of the AI text detection task. The data construction process is presented in Section \ref{subsec3_1}. To gain an overview of the dataset, we conducted dataset analysis in Section \ref{subsec3_2}.

\subsection{Dataset creation}\label{subsec3_1}

\begin{figure*}[!ht]
  \centering
  \includegraphics[width=1\textwidth]{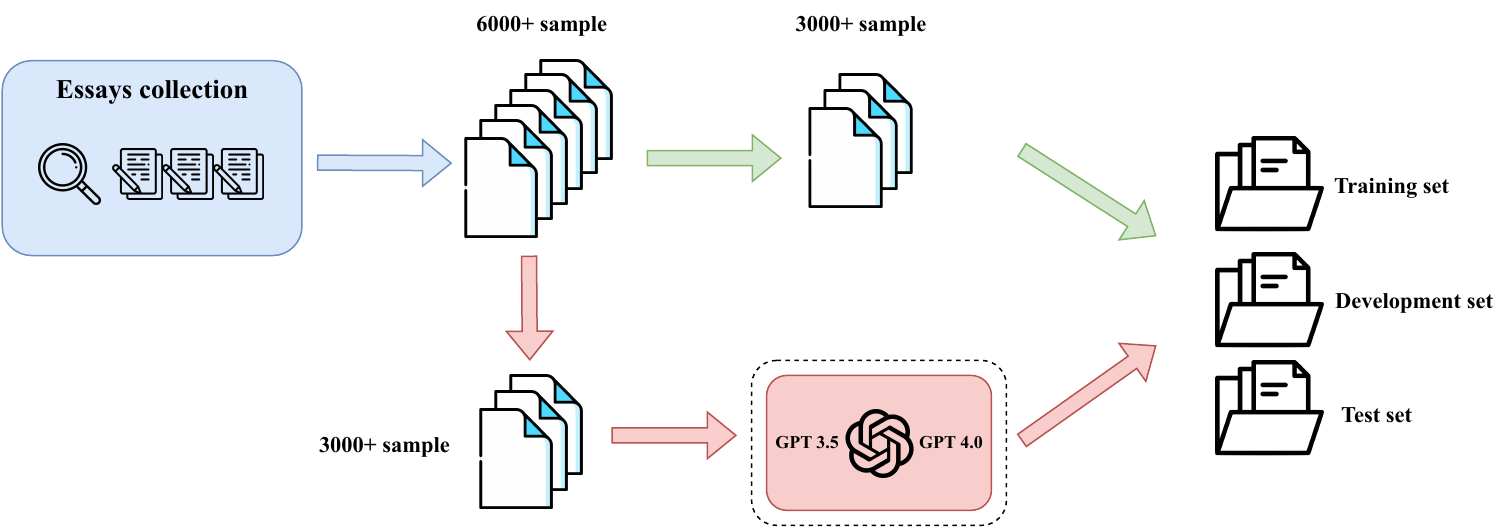}
  \caption{Overview of ViDetect dataset creation process.}
  \label{data_creation}
\end{figure*}

The dataset has been categorized into two primary labels to distinguish between texts generated by AI and texts written by humans. The data collection process focused on extracting content from educational platforms in Vietnam, containing sample essays from students across different grade levels. To ensure the authenticity of the collected texts, specific criteria were set: only texts published before 2021 were considered, before advanced text generation tools like ChatGPT became widely employed.

To guarantee a diverse representation of human writing styles, over 6,000 essays, each with a length of 256 words or more, were meticulously selected. This length requirement aimed to capture the intricacies and variations inherent in human expression. In a subsequent step, a random subset comprising 3,000 samples was chosen from the dataset. These samples were subjected to rewrites by both ChatGPT 3.5 and ChatGPT 4.0. The objective was to retain the core essence of each essay while allowing the AI models to reinterpret them. Following this, the reworked samples were automatically labeled as "1," signifying text generated by artificial intelligence. The remaining approximately 3,000 samples retained their original label of "0," denoting human-written texts.
Finally, we divided the data set into three sets training, development, and test set with a ratio of 7:1:2 to serve the process of building a deep learning model, with the goal of recognizing generated text. created by artificial intelligence.

\subsection{Dataset Analysis}\label{subsec3_2}

After data collection and preprocessing, we have obtained a complete dataset to serve the AI-generated text detection problem. Based on the details of the dataset presented in Table \ref{Table 1}, we see that AI is not capable of expressing longer paragraphs to express emotions as well as human expressions, so most of the texts written by AI creation is only a relatively short length compared to human writing.
\begin{table}[!ht]
\caption{Details about the dataset}
\centering
\begin{tabular}{ccc}
\hline
                     & \textbf{Human Writing} & \textbf{AI generated} \\ \hline
\textbf{Size}        & 3388                    & 3388                  \\ \hline
\textbf{Avg. Length} & 323                     & 301                   \\ \hline
\textbf{Max Length}  & 13112                   & 1000                  \\ \hline
\textbf{Min Length}  & 200                     & 47                    \\ \hline
\end{tabular}
\label{Table 1}
\end{table}

When we examine the average number of sentences and paragraphs in Table \ref{Table 2}, it becomes evident that AI-generated texts are typically analyzed individually, resulting in longer paragraphs. In contrast, human-written texts exhibit a higher average number of sentences, as humans tend to compose fewer paragraphs but thoroughly analyze the content from various perspectives.

\begin{table}[!ht]
\centering
\caption{Statistics on number of sentences and number of paragraphs}
\begin{tabular}{lccc}
\hline
                                                       &                          & \textbf{Human Writting} & \textbf{AI generated} \\ \hline
\multicolumn{1}{c}{\multirow{2}{*}{\textbf{Sentence}}} & \textbf{Avg}  & 13.69                      & 12.18                    \\
\multicolumn{1}{c}{}                                   & \textbf{Min}     & 3                   & 3                 \\
                                & \textbf{Max}     & 17                   & 16                 \\ \hline
\multirow{2}{*}{\textbf{Paragraph}}                    & \textbf{Avg} & 2.71                       & 3.84                     \\
                                                       & \textbf{Min}     & 1                  & 2                \\ 
                                                       & \textbf{Max}     & 8                  & 9                \\ \hline
\end{tabular}

\label{Table 2}
\end{table}

\section{Methodology}\label{sec4}

\subsection{Evaluation Metrics for LLM-generated Text Detection}

In this section, we enumerate and discuss metrics conventionally utilized in the tasks of LLM-generated text detection. These metrics include Accuracy, F1 Score, and Area Under the Receiver Operating Characteristic Curve (AUROC). Furthermore, we discuss the advantages and drawbacks associated with each metric to facilitate informed metric selection for varied research scenarios in subsequent studies.

The confusion matrix can help effectively evaluate the performance of the classification task and describes all possible results (four types in total) of the LLM-generated text detection task:

\begin{itemize}[label=--]
  \item \textbf{True Positive (TP)} refers to the result of the positive category (LLM-generated text) correctly classified by the model.
  \item \textbf{True Negative (TN)} refers to the result of the negative category (human-written text) correctly classified by the model.
  \item \textbf{False Positive (FP)} refers to the result of the positive category (LLM-generated text) incorrectly classified by the model.
  \item \textbf{False Negative (FN)} refers to the result of the negative category (human-written text) incorrectly predicted by the model.
\end{itemize}

\subsubsection{Accuracy}

Accuracy serves as a general metric, denoting the ratio of correctly classified texts to the total text count. While suitable for balanced datasets, its utility diminishes for unbalanced ones due to sensitivity to category imbalance. 

Accuracy can be described by the following formula:

\begin{equation}
\text{Accuracy} = \frac{\text{correctly detected samples}}{\text{all samples}} = \frac{TP + TN}{TP + TN + FP + FN}
\end{equation}








\subsubsection{F1 Score}

The F1 Score constitutes a harmonic mean of precision and recall, integrating considerations of false positives and false negatives. It emerges as a prudent choice when a balance between precision and recall is imperative. The F1 score can be calculated using the following formula:

\begin{equation}
\text{F1} = 2 \cdot \frac{\text{Precision} \cdot \text{Recall}}{\text{Precision} + \text{Recall}} = \frac{2 \cdot TP}{2 \cdot TP + FP + FN}
\end{equation}

\subsubsection{AUROC}

The AUROC metric, originating from Receiver Operating Characteristic (ROC) curves, evaluates classification performance across various thresholds by examining true positive and false positive rates. It is valuable for assessing classification effectiveness under different thresholds, which is particularly important in scenarios with imbalanced datasets and binary classification tasks where specific false positive and miss rates are of concern. The AUROC calculation formula is as follows:

\begin{equation}
\text{AUROC} = \int_{0}^{1} \frac{TP}{TP + FP} \,d\left(\frac{FP}{FP + TN}\right)
\end{equation}
\subsection{Data Pre-processing}
To enhance the quality and relevance of the dataset, thereby optimizing the performance and interpretability of the AI text detection model during training and evaluation, we applied several preprocessing techniques as detailed below:
\begin{enumerate}
    \item \textbf{Removing unnecessary characters}: We remove punctuation, special characters, numbers, or any other symbols that have no linguistic meaning. By taking this measure, the dataset was refined to contain only essential text elements relevant to the task of language comprehension.
    \item \textbf{Removing stop words}: We removed common words in the Vietnamese language and words that appear a lot in the dataset but these words do not have important meanings.
    
  \item \textbf{Lowercasing}: This procedure served the primary purpose of standardizing the textual data, ensuring consistency and mitigating unnecessary duplications. By converting all text to lowercase, potential discrepancies in casing were mitigated, fostering a more coherent and streamlined dataset for the subsequent modeling phase.
\end{enumerate}

\subsection{Baseline models}
To match Vietnamese and at the same time improve performance on the dataset, we followed the fine-tuning LMs methods as Section \ref{fine-tuning} and chose pre-trained language models that not only reflect  the richness of the language but also help to improve the ability to process and deeply understand text in the Vietnamese context as follows:

\textbf{ViT5-base}: Vietnamese Text-to-Text Transformer \cite{phan2022vit5} is a pre-trained Transformer-based encoder-decoder model for the Vietnamese language. It was developed by VietAI and is trained on a massive corpus of high-quality and diverse Vietnamese texts using T5-style self-supervised pretraining.

\textbf{BARTpho-base}: BARTpho \cite{tran2021bartpho} is a pre-trained sequence-to-sequence model for Vietnamese developed by VinAI. It is based on the BART architecture \cite{lewis2019bart}, which is a denoising autoencoder model that has been shown to be effective for a variety of natural language processing tasks. BARTpho is trained on a massive corpus of Vietnamese text data using the BART pre-training scheme.

\textbf{PhoBERT-base}: PhoBERT \cite{nguyen2020phobert} is a state-of-the-art monolingual pre-trained language model for Vietnamese. It was developed by VinAI Research and is trained on a massive corpus of Vietnamese text data using the RoBERTa \cite{liu2019roberta} pre-training approach. PhoBERT outperforms previous monolingual and multilingual approaches, obtaining new state-of-the-art performances.

\textbf{mDeBERTa-v3-base}: mDeBERTa-v3-base \cite{he2022debertav3} is one of state-of-the-art multilingual pre-trained Transformer-based encoder-decoder model developed by Microsoft. It is a multilingual version of DeBERTa \cite{he2020deberta}, which was trained on a massive corpus of 2.5T CC100 data using ELECTRA-Style pre-training with Gradient Disentangled Embedding Sharing. This training method allows mDeBERTa-v3-base to effectively capture the nuances of many languages and perform many subsequent tasks well in many languages including Vietnamese.

\textbf{BERT multilingual base cased}: BERT multilingual base cased \cite{pires2019multilingual} is a variant of BERT \cite{devlin2018bert}, this is a pre-trained multilingual Transformer-based masked language modeling (MLM) model developed by Google AI. It is a cased model, which means that it distinguishes between different cases of letters, such as uppercase and lowercase. The model is pre-trained in 104 different languages, including Vietnamese. This training allows the model to capture the nuances of multiple languages and perform well on a variety of downstream tasks in multiple languages.

\section{Experiment and Results}\label{sec5}

\subsection{Experimental Configuration}
All baseline models were trained and fine-tuned using the Adam optimization. We utilized an RTX 4090 setup with 24GB of memory to fine-tune models. The hyperparameters for the models are set as follows: learning rate = 3e-05, dropout = 0.2, batch size = 4, and early stopping patience = 5.
\subsection{Experimental Results}
In the context of AI-generated text detection, our study evaluates the performance accuracy of several baseline models across different token lengths. The models considered include ViT5-base, BARTpho-syllable-base, PhoBERT-base, mDeBERTa-v3-base, and BERT-multilingual-base-cased, with assessments conducted at token lengths of 64, 128, and 256.
\subsubsection{Performance Accuracy for Baseline Models}

To provide a comprehensive overview, we calculate the average accuracy across all models and token lengths. The average accuracy is recorded as 0.8580 for 64 tokens, 0.8721 for 128 tokens, and 0.8437 for 256 tokens. In this metric, experiments show that increasing the number of tokens does not make the model better in terms of overall results. More details are in Table \ref{tab:baseline-acc}

\begin{table}[!ht]
\caption{Performance Accuracy for Baseline Models}
\label{tab:baseline-acc}
\begin{tabular}{lccc}
\toprule
\textbf{Baseline Models} & \multicolumn{3}{c}{\textbf{Accuracy}}  \\
\cmidrule(lr){2-4}
                         & 64 tokens & 128 tokens & 256 tokens \\ \midrule
ViT5-base                & 0.8611& 0.8438& 0.8591\\
BARTpho-syllable-base    & 0.8570& 0.8840& 0.8632\\
PhoBERT-base             & 0.8743& 0.8819& 0.8708\\
mDeBERTa-v3-base         & 0.8957& 0.8542& 0.7975\\
BERT-multilingual-base-cased & 0.8017& 0.8964& 0.8279\\ \midrule
\textbf{Avg}              & 0.8580& 0.8721& 0.8437\\
\bottomrule
\end{tabular}
\end{table}

\subsubsection{Performance F1 for Baseline Models}

The results from Table \ref{tab:baseline-f1} show that with the average F1 score, serving as a comprehensive performance metric, being 0.9252, 0.8897, and 0.8729 for token lengths of 64, 128, and 256 respectively. This aggregate measure provides an overall assessment of the baseline models, facilitating comparison of their overall effectiveness across different token lengths. Similar to accuracy, in these figures, the experiment demonstrates that increasing the number of tokens does not improve the model's performance.

\begin{table}[!ht]
\caption{Performance F1 for Baseline Models}
\label{tab:baseline-f1}
\begin{tabular}{lccc}
\toprule
\textbf{Baseline Models} & \multicolumn{3}{c}{\textbf{F1}}  \\
\cmidrule(lr){2-4}
                         & 64 tokens & 128 tokens & 256 tokens \\ \midrule
ViT5-base                & 0.9041     & 0.8551      & 0.8579      \\
BARTpho-syllable-base    & 0.9016     & 0.9170      & 0.8852       \\
PhoBERT-base             & 0.9164     & 0.9301      & 0.8893      \\
mDeBERTa-v3-base         & 0.9506     & 0.8955      & 0.8447      \\
BERT-multilingual-base-cased & 0.9532  & 0.8509      & 0.8876       \\ \midrule
\textbf{Avg}              & 0.9252    & 0.8897     & 0.8729     \\
\bottomrule
\end{tabular}
\end{table}

\subsubsection{Performance AUROC for Baseline Models}
The table \ref{tab:baseline-auroc} presents the AUROC performance results of the baseline models on three different input data sizes (64 tokens, 128 tokens and 256 tokens). The average of all models (Avg) shows a significant improvement from 0.8629 (64 tokens) to 0.9168 (256 tokens). This suggests that, overall, models tend to improve performance with more input. Unlike Accuracy, F1, experiments show that increasing the number of tokens also improves model performance.
\begin{table}[!ht]
\caption{Performance AUROC for Baseline Models}
\label{tab:baseline-auroc}
\begin{tabular}{lccc}
\toprule
\textbf{Baseline Models} & \multicolumn{3}{c}{\textbf{AUROC}}  \\
\cmidrule(lr){2-4}
                         & 64 tokens & 128 tokens & 256 tokens \\ \midrule
ViT5-base                & 0.8879     & 0.9101      & 0.9404      \\
BARTpho-syllable-base    & 0.8817     & 0.9239      & 0.9180       \\
PhoBERT-base             & 0.9013     & 0.8989      & 0.9304      \\
mDeBERTa-v3-base         & 0.8211     & 0.8837      & 0.8991      \\
BERT-multilingual-base-cased & 0.8227  & 0.8338      & 0.8960       \\ \midrule
\textbf{Avg}              & 0.8629    & 0.8901     & 0.9168     \\
\bottomrule
\end{tabular}
\end{table}

\section{Conclusion and Future Works}\label{sec6}

In summary, our study significantly enhances the capability to detect Large Language Model (LLM)-generated text in the Vietnamese language, underscoring the necessity for language-specific strategies. Utilizing a \textbf{ViDetect} dataset comprising 6.800 samples, obtained through a combination of self-crawling and prompting from the AI tool GPT chat, along with the application of advanced methods like ViT5, BARTpho, Mderbertav3, PhoBERT and Bert-multilingual, our research highlights the adaptability and effectiveness of diverse techniques within the Vietnamese context.

Looking ahead, future efforts should delve into exploring multimodal approaches, refining models for specific domains, conducting adversarial testing, integrating user feedback, and considering ethical implications to further improve model development and address emerging challenges.

\bibliography{sn-bibliography}

\end{document}